\documentclass[5pt]{article}

\usepackage[letterpaper]{geometry}
\usepackage{spconf,amsmath,epsfig}
\usepackage{enumitem}

\usepackage{fancyhdr}
\usepackage{graphicx}
\usepackage{amsfonts}
\usepackage{amsmath}
\usepackage{amsthm}
\usepackage{varwidth}
\usepackage{algorithm}
\usepackage{algorithmic}
\usepackage{subfigure}
\usepackage[numbers]{natbib}

\usepackage{caption}
\usepackage{setspace}

\begin{document}\sloppy
\topmargin=0mm

\def\x{{\mathbf x}}
\def\L{{\cal L}}

\newtheorem{thm}{Theorem} 

\def\ie{\emph{i.e}.} 
\def\eg{\emph{e.g}. } 
\def\etal{\emph{et al}. } 
\def\st{\emph{s.t}. } 

\title{Continuity-Discrimination Convolutional Neural Network
\center{for Visual Object Tracking}}
%
%
%

\name{Shen Li$^{\dagger, \star}$ \quad Bingpeng Ma$^{\star}$ \quad Hong Chang$^{\dagger}$ \quad Shiguang Shan$^{\dagger, \star}$ \quad Xilin Chen$^{\dagger}$}
\address{
	${\dagger}$ Key Lab of Intelligent Information Processing of Chinese Academy of Sciences (CAS),\\
	Institute of Computing Technology, CAS, Bejing, 100190, China.\\
	$^{\star}$ School of Computer and Control Engineering, \\
	University of Chinese Academy of Sciences, Beijing 100049, China.\\
	\{shen.li, bingpeng.ma, hong.chang, shiguang.shan, xilin.chen\}@vipl.ict.ac.cn}

\maketitle

\begin{abstract}
This paper proposes a novel model, named Continuity-Discrimination Convolutional Neural Network (CD-CNN), for visual object tracking. 
Existing state-of-the-art tracking methods do not deal with temporal relationship in video sequences, which leads to imperfect feature representations. 
To address this problem, CD-CNN models temporal appearance continuity based on the idea of temporal slowness. 
Mathematically, we prove that, by introducing temporal appearance continuity into tracking, the upper bound of target appearance representation error can be sufficiently small with high probability. 
Further, in order to alleviate inaccurate target localization and drifting, we propose a novel notion, object-centroid, to characterize not only objectness but also the relative position of the target within a given patch. 
Both temporal appearance continuity and object-centroid are jointly learned during offline training and then transferred for online tracking. 
We evaluate our tracker through extensive experiments on two challenging benchmarks and show its competitive tracking performance compared with state-of-the-art trackers.
\end{abstract}


\section{Introduction}
\label{sec:intro}
Visual object tracking is one of the most fundamental topics in computer vision with a wide range of multimedia applications, including surveillance, vehicle navigation, human computer interaction and so forth. 
Various relevant approaches have been proposed over the past several decades \citep{Li:2013:SAM:2508037.2508039, 6671560}. 
Most of them focus on learning target appearance representations that are robust to various disturbing factors. 
Particularly, some tracking methods \citep{bertinetto2016fully, tao2016siamese} leverage high-level CNN features for more robust representations of target appearance. 
These methods learn either a similarity matching function explicitly or implicitly, or a discriminative classifier offline for online tracking within the tracking-by-detection framework. 


However, the deep learning based methods above 
do not deal with temporal relationship between video frames. 
Visual tracking is inherently a temporal problem. Posing tracking as a totally frame-unrelated learning task leads to loss of temporal information in appearance representations, which cannot benefit the robustness of features under disturbing conditions. Therefore, it is essential to develop a learning method to properly deal with temporal information of target representations.

Besides, drifting is still a challenging problem that remains unresolved in visual tracking. This problem depends largely on the lack of discriminability of objectness. From a practical point of view, a target to be tracked is basically a semantic object. Hence, within tracking-by-detection framework, a proper definition of objectness can be exploited as a criterion to filter most of distracting candidates of non-objects. Wang \etal \citep{wang2015transferring} introduced the \emph{objectness} concept into tracking and designed a CNN that maps from an image patch into a probability map. Each pixel in the probability map indicates the probability of being part of an object. 
However, in online tracking, their method resorts to an exhaustive search that includes an inverse mapping from the probability map to the raw image. This coarse inverse mapping may cause inaccurate target localization especially in face of fast motion and background clutter.

In this paper, we propose continuity-discrimination convolutional neural network (CD-CNN) to jointly model temporal continuity and objectness discrimination for visual object tracking. 
Firstly, we utilize temporal slowness principle \citep{cogprints2804} to learn temporally continuous feature representations that are robust to varying object appearances and environments. 
Temporal slowness, according to recent discoveries in cognitive science, is considered as a possible learning principle of complex cells in primary visual cortex. 
Slowly varying features extracted from even the quickly varying signals can possibly reflect the inherent properties of the environment and thus are robust to frequently intensive transformations.
In cognitive science and computer vision areas, various related researches have been carried out recently. 
Li \etal \citep{li_unsupervised_2008} reported that unsupervised temporal slowness learning enables the responses of neurons to be selective to different objects, yet tolerant or invariant to changes in object position, scale and pose.
Zou \etal \citep{Zou2012DeepLO} proposed a hierarchical network to learn invariant features under temporal slowness constraints. The resulting spatial feature representation is well suited for the still image classification problem. 
In our method, we penalize the temporal discontinuity of feature representations using $\ell_2$-norm. In this way, the learned features can characterize temporal invariance of target appearance. Then, the learned temporal appearance continuity is transferred to a specific target in online tracking. Empirically, such transferring helps improve the tracking performance of our tracker. And mathematically, with temporal appearance continuity introduced into tracking, the upper bound of target appearance representation error can be sufficiently small with high probability.

Secondly, we introduce a novel notion, {\em object-centroid}, to better describe the discriminability of target against backgrounds. Compared with objectness, object-centroid characterizes not only the object per se but also the relative position of a target in a given candidate patch. 
Within the tracking-by-detection framework, object-centroid can be utilized to directly distinguish a target from backgrounds without extra strategies for bounding box determination. To this end, we design a particular sampling protocol to draw object-centroid training samples and feed them into our deep network for discrimination learning. Consequently, in experiments, our tracker is more sensitive to candidate patches of object-centroid and hence drifting is alleviated.

\section{Proposed Method}
\label{sec:proposed_method}
In this section, we first define temporal appearance continuity mathematically and present the notion of object-centroid. Then, the proposed CD-CNN architecture for offline training is illustrated and so is the online tracking algorithm.

\subsection{Temporal Appearance Continuity}

In general, tracking performance highly depends upon the robustness of target appearance representation. Zou \etal \citep{Zou2012DeepLO} has pointed out that with temporal slowness constraints, the learned features manifest robustness and even invariance to challenging variations. Inspired by their views, we mathematically define temporal appearance continuity of the target between frames based on temporal slowness.
By doing so, we make use of inherent temporal continuity to learn appearance representations that are robust to variations.


Formally, given a feature mapping $\Phi: \mathbb{R}^r \mapsto \chi \subset \mathbb{R}^n$, where $r$ is the dimensionality of the raw pixel space and $\chi$ is the feature space, the temporal appearance continuity of a target patch is formalized by Lipchitz continuity of $\Phi$ at any dimension with respect to time $t$:
\begin{equation}
{\left\vert \Phi_i\left(P_*^{(t+\Delta t)}\right)-\Phi_i\left(P_*^{(t)}\right) \right\vert} \le K{\Delta t}, \forall i = 1, ..., n,
\end{equation}
where $\Phi_i(P_*^{(t)})$ is the $i$th component of the $n$-dimensional feature vector of the ground-truth patch $P_*^{(t)}$ in the $t$th frame and $K$ is a positive constant.  
We use Euclidean norm to measure the distance between features. As $\Delta t \to 0$, we have
\begin{equation}
\begin{split}
& 	{\left\lVert \Phi\left(P_*^{(t+\Delta t)}\right)-\Phi\left(P_*^{(t)}\right) \right\rVert}_2 \\
& \le {\left\lVert \Phi\left(P_*^{(t+\Delta t)}\right)-\Phi\left(P_*^{(t)}\right) \right\rVert}_1 
 < nK\Delta t \to 0.
\end{split}
\end{equation}

Consequently, if the frame switch of a given video sequence is sufficiently smooth, the Euclidean distance between high-level features is bounded by some small $\epsilon:=nK\Delta t$.

\subsection{Object-centroid Discrimination}
\begin{figure}[ht]
\centering
\includegraphics[scale=0.4]{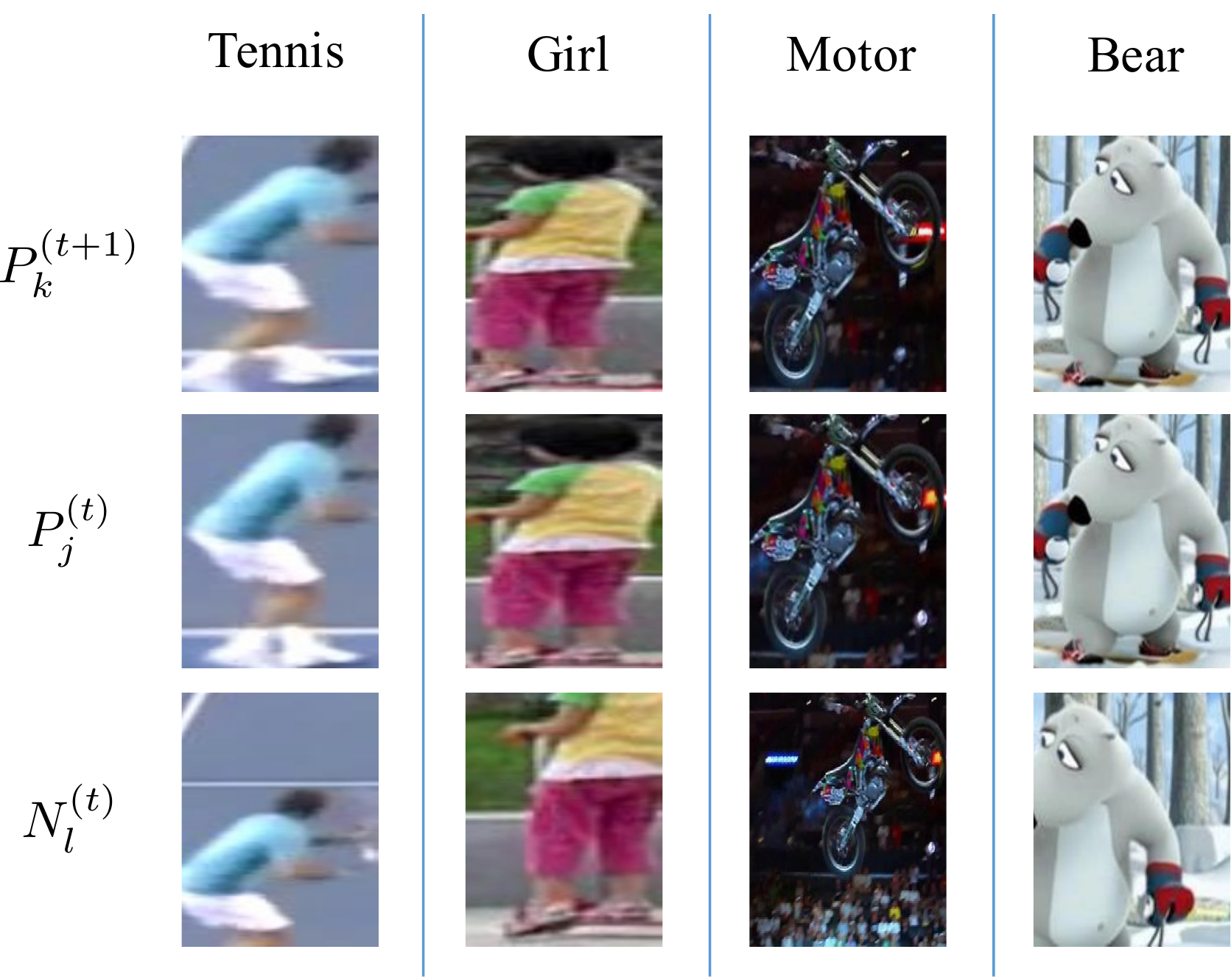}
\caption{Training examples for object-centroid discrimination learning. $P_j^{(t)}$ and $N_l^{(t)}$ denote the $j$th positive sample and the $l$th negative sample drawn from the $t$th frame, respectively, and $P_k^{(t+1)}$ denotes the $k$th positive sample drawn from the $(t+1)$th frame. 
Positive samples have bounding boxes that tightly envelop the entire target while negative ones either include more background or just partial target. Note that the only difference between $P_j^{(t)}$ and $N_l^{(t)}$ drawn from Motor sequence is that $N_l^{(t)}$ includes redundant background while $P_j^{(t)}$ tightly envelops the motorbike. This difference is non-trivial, since such negative samples highlight the loss of {\em positional centroid}.
}
\label{fig:pos_neg}
\end{figure}

Within the Bayesian tracking framework, candidate bounding boxes are drawn from a certain probability distribution in the online tracking phase. 
To obtain accurate target localization, our aim is to select the candidate with the whole target located in the center of the bounding box and as less background left as possible. Apparently, such selection depends largely on specific training data with which a discriminative classifier is trained. A candidate patch that contains object semantics is of objectness. However, satisfying objectness is not sufficient for accurate target localization. It is necessary to search for a candidate that tightly envelops the entire target, leaving it located in the center of the patch (positional centroid). Those that are of objectness and positional centroid are literally defined to satisfy {\em object-centroid}. Figure \ref{fig:pos_neg} illustrates some positive samples that satisfy object-centroid and negative samples with loss of object-centroid. 
These properties can be exploited to train a discriminative binary classifier so as to filter distracting samples that contain redundant background or partial target. 

Specifically, training data for object-centroid discrimination are generated using a particular sampling protocol: the positive samples are one or two pixels shifted from the groundtruths, and the negative are randomly drawn from a distribution under the constraint that the intersection-over-union (IoU) between a negative sample and its corresponding groundtruth is subject to $0 < lo < \text{IoU} \le hi$, for some pre-specified $lo$ and $hi$. Note that the non-zero lower threshold $lo$ is an essential factor to avoid object-centroid inconsistency. For example, a sample of zero IoU with the corresponding ground-truth might also satisfy the object-centroid property (i.e., it tightly contains similar distracting target in its center), thus should be considered as a positive sample instead. 




\subsection{Offline Training}
Suppose the training set is given by $\mathbb{S}=\{P_j^{(t)}, N_l^{(t)} \vert j=1,...,m_p^{(t)}, l=1,...,m_n^{(t)}, t=1,...,T\}$, 
where $P_j^{(t)}$ is the $j$th positive sample satisfying the object-centroid property and $N_l^{(t)}$ is the $l$th negative sample drawn from the $t$th frame. 
Before being fed into CD-CNN, all samples are resized into 224-by-224 and normalized.
In the offline training stage, temporal appearance continuity and object-centroid discrimination are jointly utilized to learn a deep discriminative model for generic object tracking. 
To this end, we design Continuity-Discrimination CNN (CD-CNN), in which temporal appearance continuity and object-centroid are formulated as specific loss functions, respectively.

More specifically, with $\Delta t$ set to 1 (unit time), the loss function corresponding to temporal appearance continuity is defined as, for all $j, k, t$,
\begin{equation}
\label{eq:c_loss}
\mathcal{L}^{C}\left(P_j^{(t)}, P_k^{(t+1)}\right) = {\left\lVert \Phi\left(P_j^{(t)}\right)-\Phi\left(P_k^{(t+1)}\right) \right\rVert}_2^2,
\end{equation}
where $\Phi(\cdot)$ is the nonlinear mapping from the sample space to the feature space $\chi$.

To encourage object-centroid discriminability, we define the following loss function, for $\beta > 0$ and all $j, l, t$,
\begin{equation}
\label{eq:d_loss}
\mathcal{L}^{D}\left(P_j^{(t)}, N_l^{(t)}\right) = \exp\left(-\beta{\left\lVert \Phi\left(P_j^{(t)}\right)-\Phi\left(N_l^{(t)}\right) \right\rVert}_2^2\right).
\end{equation}

In addition, to directly quantify object-centroid, the output of $\Phi(\cdot)$ is fed into a nonlinear binary classifier which indicates the probability of being object-centroid, $p(s)$, for all $s\in \mathbb{S}$. We define the binary classification loss as the soft-max cross entropy loss, for all $j, l, t$,
\begin{equation}
\label{eq:s_loss}
\mathcal{L}^{S}  \left(P_j^{(t)}, N_l^{(t)}\right) = -\log\left(1-p\left(N_l^{(t)}\right)\right)p\Bigl(P_j^{(t)}\Bigr).
\end{equation}
Formally, the optimization objective of the offline joint learning task is formulated as
\begin{equation}
\label{eq:min_total_loss}
\min_{W} \sum_{t=1}^T \sum_{j,k,l} \mathcal{L}\left(P_j^{(t)}, P_k^{(t+1)}, N_l^{(t)}\right)
\end{equation}
where
\begin{equation}
\begin{split}
\label{eq:total_loss}
\mathcal{L}\left(P_j^{(t)}, P_k^{(t+1)}, N_l^{(t)}\right) 
& = \mathcal{L}^{C}\left(\cdot, \cdot\right) + \left(\lambda \mathcal{L}^{D}+\mu \mathcal{L}^{S}\right) \left(\cdot, \cdot\right).
\end{split}
\end{equation}
Here, $\lambda > 0$ and $\mu > 0$ are trade-off coefficients, and $W$ denotes all the learnable parameters.

\begin{figure}[t] 
\centering
\includegraphics[width=1\linewidth]{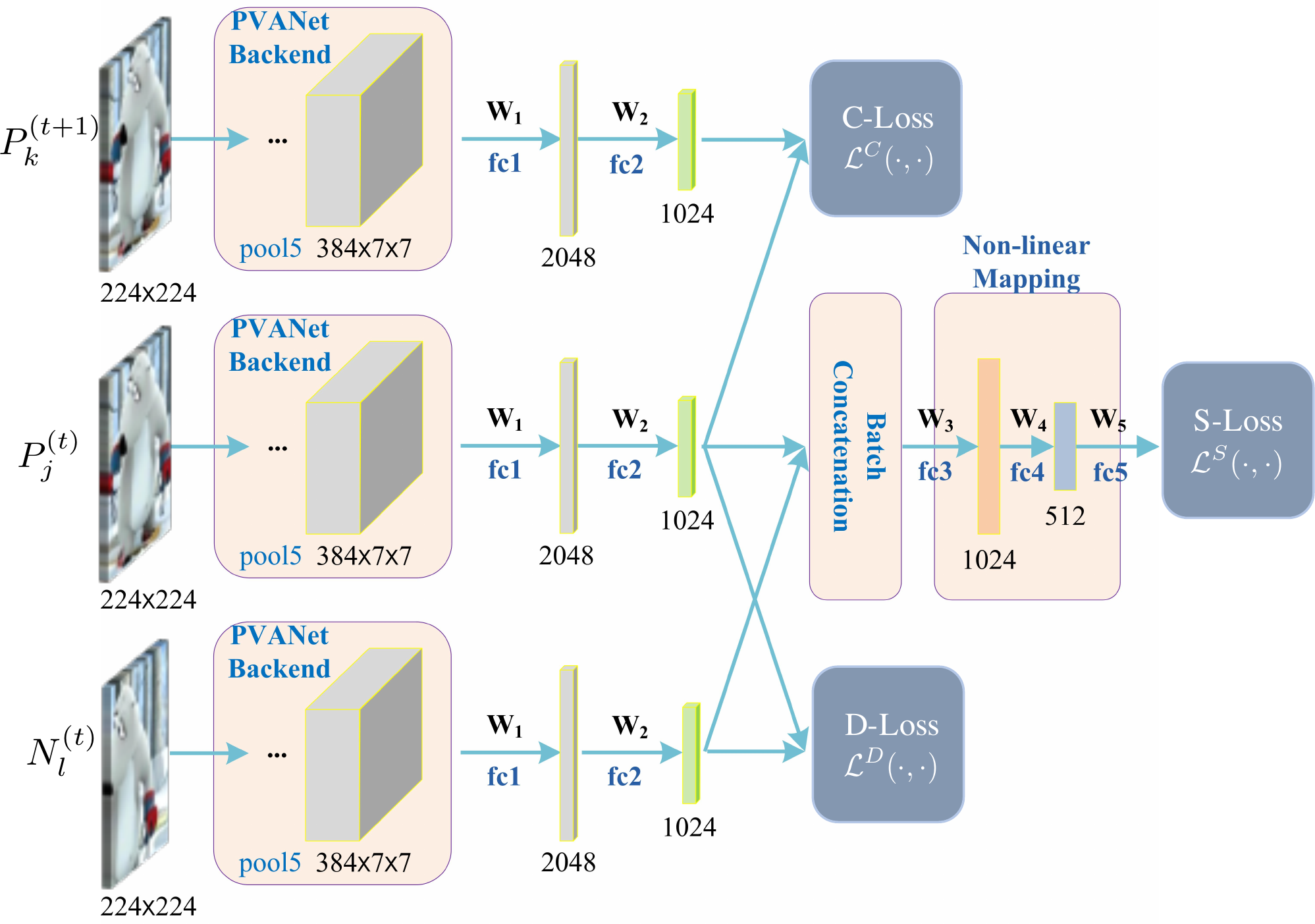}
\caption{CD CNN Structure for Offline Joint training}
\label{fig:network_structure}
\end{figure}

In practice, we choose PVANET \citep{kim2016pvanet} as the backend of CD-CNN. The overall architecture of our network is illustrated in Figure \ref{fig:network_structure}. 
As is shown in Figure \ref{fig:network_structure}, the network takes a triplet of RGB image patches ${P_j^{(t)}, P_k^{(t+1)},N_l^{(t)}}$ as input. On top of PVANET pool5, two fully-connected layers (fc1 and fc2) with learnable parameters $\{W_1, W_2\}$ are added in form of a nonlinear mapping from the sample space to the feature space $\chi$. 
Besides, after batch concatenation, features are fed into the nonlinear binary classifier in form of three fully-connected layers (fc3, fc4 and fc5) with learnable parameters $\{W_3, W_4, W_5\}$.
In experiments, we will show the necessity of all these loss functions \eqref{eq:c_loss}\eqref{eq:d_loss}\eqref{eq:s_loss} as basic components of the overall loss function \eqref{eq:total_loss}.

\subsection{Online Tracking}
\label{sec:online_tracking}
The overall online tracking algorithm is summarized in Algorithm 1 in the Appendix A. 
Through offline training, our CD-CNN has learned temporal appearance continuity for generic objects. Then, in online tracking, this property can be transferred to any specific object to be tracked. 
In order to focus on a single specific target, it is necessary to finetune the network with the initial testing frame. Specifically, all model parameters are updated by minimizing Eqn.~\eqref{eq:min_total_loss} as in offline training. The only difference is that the temporal appearance continuity loss, Eqn.~\eqref{eq:c_loss}, is replaced by the Euclidean distance in $\chi$ between positive samples in the same frame, i.e., $\mathcal{L}^{P}(P_j^{(t)}, P_k^{(t)})= {\lVert \Phi(P_j^{(t)})-\Phi(P_k^{(t)}) \rVert}_2^2$. In this way, we can not only maintain the transferred continuity but also ensure similar feature representations for positive samples.

After finetuning CD-CNN, for each subsequent frame of the given tracking sequence, our tracker draws candidates from a Gaussian distribution, performs one-pass forward for each and selects top five candidates with the highest object-centroid scores $p(\cdot)$. 
The final bounding box in the $t$th frame is determined by averaging their sizes and locations.

\subsection{Target Appearance Representation Error}
The target appearance representation error incurred by our tracker can be defined as the Euclidean distance $\mathcal{E} := {\lVert \Phi({\hat P}_*^{(t+1)})-\Phi(P_*^{(t+1)}) \rVert}_2$, where $P_*^{(t)}$ is the ground-truth and ${\hat P}_*^{(t)}$ is the predicted target patch in the $t$th frame.

\begin{thm}[Upper Bound of Target Appearance Representation Error]
With probability no less than $1-\rho$, the target appearance representation error $\mathcal{E}$ is upper-bounded by $\sum_j{\sqrt{\hat{\mathcal{L}_j^{C}}}}/m + n(\delta + K\Delta t)$, for any $\delta > \sqrt{{n \over m} max_i{\mathrm{Var}\left(\Phi_i\right)}}$, where $\hat{\mathcal{L}_j^{C}} = {\lVert \Phi(\hat{P}_*^{(t+1)}) - \Phi(P_j^{(t)}) \rVert}_2^2$ is the estimated temporal appearance continuity loss for the predicted target in the $(t+1)$th frame with respect to $P_j^{(t)}$ and $\rho = n\max_i{\mathrm{Var}(\Phi_i)}/{m\delta^2}$.
\end{thm}

The proof of Theorem 1 is presented in Appendix B. According to Theorem 1, for a tight upper bound, $\delta$ and $\Delta t$ should be $O\left({1 \over n^{1+\alpha}}\right)$ for some small $\alpha>0$. Thus, with high probability, the number of samples drawn from each frame is $m=O\left(n^{3+\alpha}\right)$. In other words, with $O\left(n^{3+\alpha}\right)$ samples, the representation error can be upper-bounded by a small value with high probability, if $\sum_j{\sqrt{\hat{\mathcal{L}_j^{C}}}}/m$ converges.

\section{Experiments}
\subsection{Implementation Details}
The training video sequences are selected from ALOV \citep{6671560}, Deform-SOT \citep{du2016online} and VOT \citep{kristan2015visual} without overlapping with the benchmarks \citep{wu2015object, Wu_2013_CVPR}. Since the ground-truths in ALOV were only annotated every five frames, we annotate ground-truths for the rest of the frames. In order to jointly learn temporal appearance continuity and object-centroid discrimination, the training data are generated from two consecutive frames with non-occluded targets that highlight object-centroid. Specifically, positive samples are generated with one or two pixels shifted from the ground-truth while negative samples are randomly drawn under the constraint of $0.2 \le \text{IoU} \le 0.6$. 

During offline training phase, we set $\lambda=\mu=10$ and $\beta=1$ and train CD-CNN for 20K iterations using ADAM. 
The initial learning rate is 0.001 for the fully connected layers (fc1$\sim$5). 
The convolutional layers are initialized by PVANET model pre-trained on ImageNet. 
In the initial frame of a test sequence, the fully-connected layers are fine-tuned for 300 iterations using SGD with learning rate 0.001. 
In each of the subsequent frames, 800 candidates ($m=800$) are sampled and evaluated. 
In handling appearance variations, we decide to update our tracking model according to the object-centroid score every five frames. The model is updated by finetuning the fc layers using samples drawn within a local window centered at the previously predicted target location. 
The local window is twice the size of the previously predicted bounding box. 
These samples are labeled according to IoU thresholds, 0.9 for the positive and 0.6 for the negative. 

\subsection{Evaluation}
\noindent {\bf Datasets and metrics:}
We empirically evaluate our proposed method on the OTB2015 Benchmark \citep{wu2015object} and the OTB2013 Benchmark \citep{Wu_2013_CVPR}. These testing sets cover various challenging conditions in visual tracking, including fast motion, deformation, background clutter and occlusion. 
For evaluation, two metrics are utilized: success plot and precision plot \citep{wu2015object}. 
Trackers are ranked according to the precision at the threshold of 20 (Prec@20) and the area-under-curve (AUC) score of the success plot.

\begin{figure*}[ht]
\centering
\includegraphics[width=0.85\linewidth]{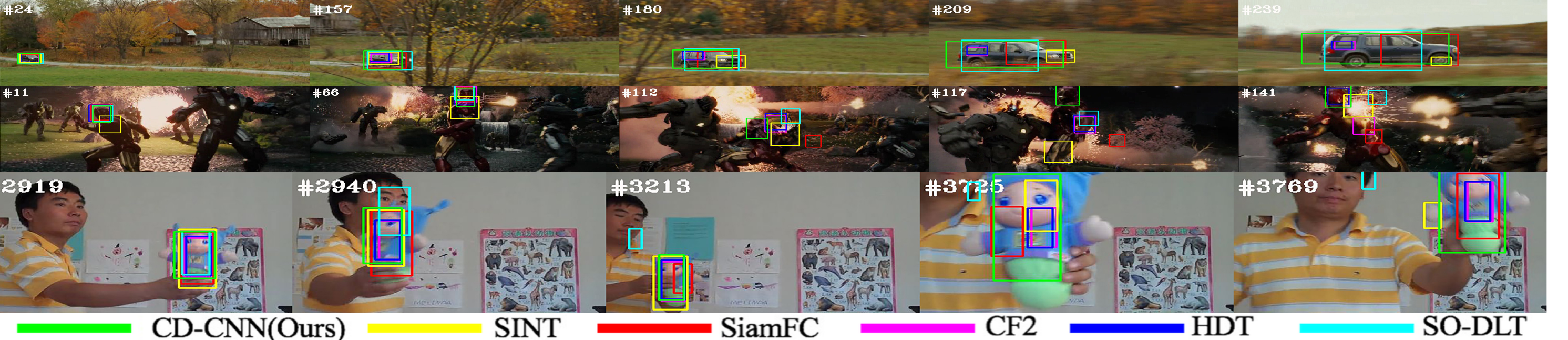}
\caption{Qualitative comparison on some challenging sequences including CarScale, Ironman and Doll.}
\label{fig:qual_perf}
\end{figure*}

\begin{figure}[ht]
	\centering
	\begin{varwidth}[ht]{\textwidth}
		\includegraphics[width=0.23\linewidth]{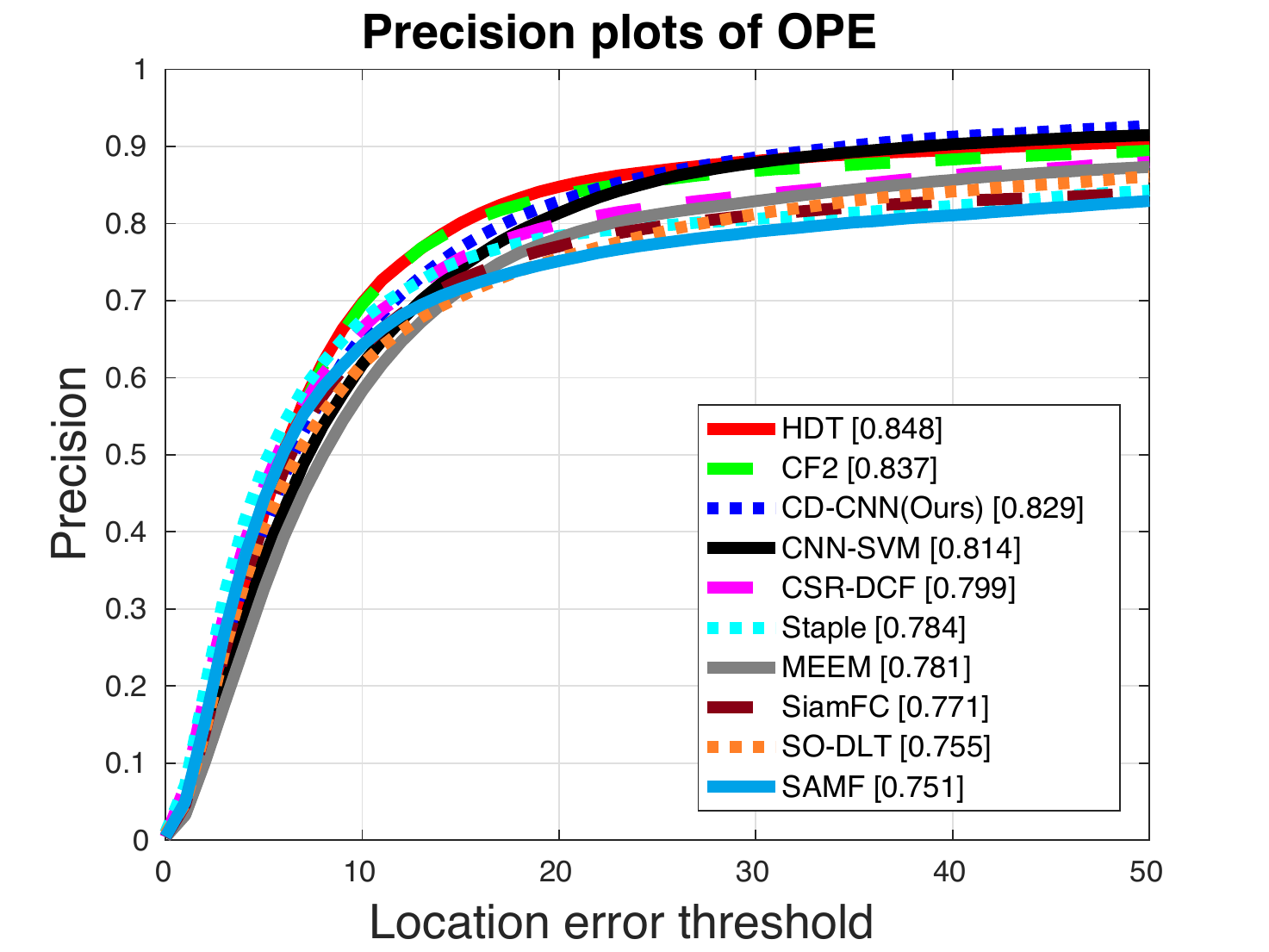}
	\end{varwidth}
	\begin{varwidth}[ht]{\textwidth}
		\includegraphics[width=0.23\linewidth]{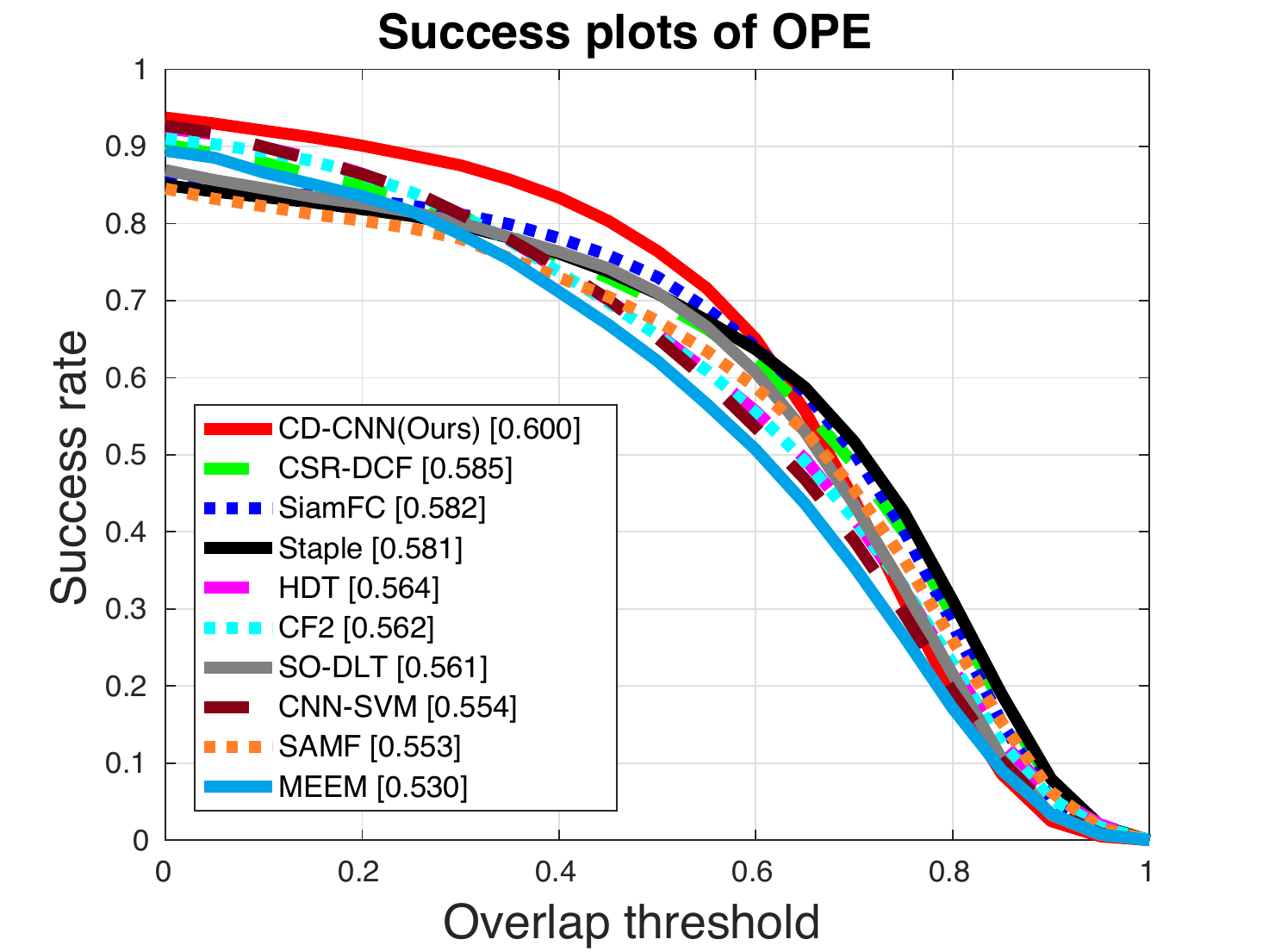}
	\end{varwidth}
	\caption{Overall performance comparison on OTB2015.}
	\label{fig:all_comp_OTB100}
\end{figure}
\begin{figure}[ht]
	\centering
	\begin{varwidth}[ht]{\textwidth}
		\includegraphics[width=0.22\linewidth]{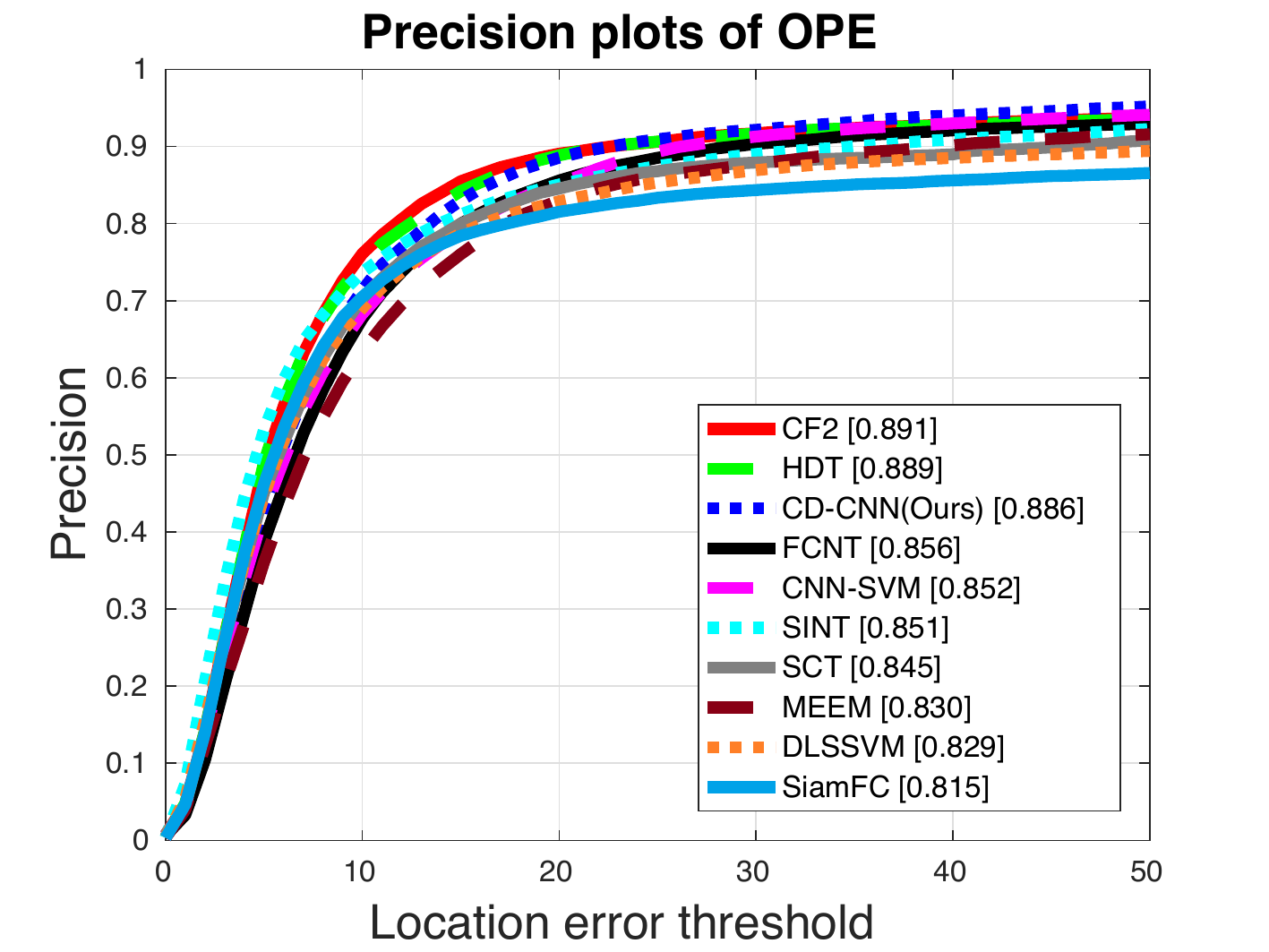}
	\end{varwidth}
	\begin{varwidth}[ht]{\textwidth}
		\includegraphics[width=0.24\linewidth]{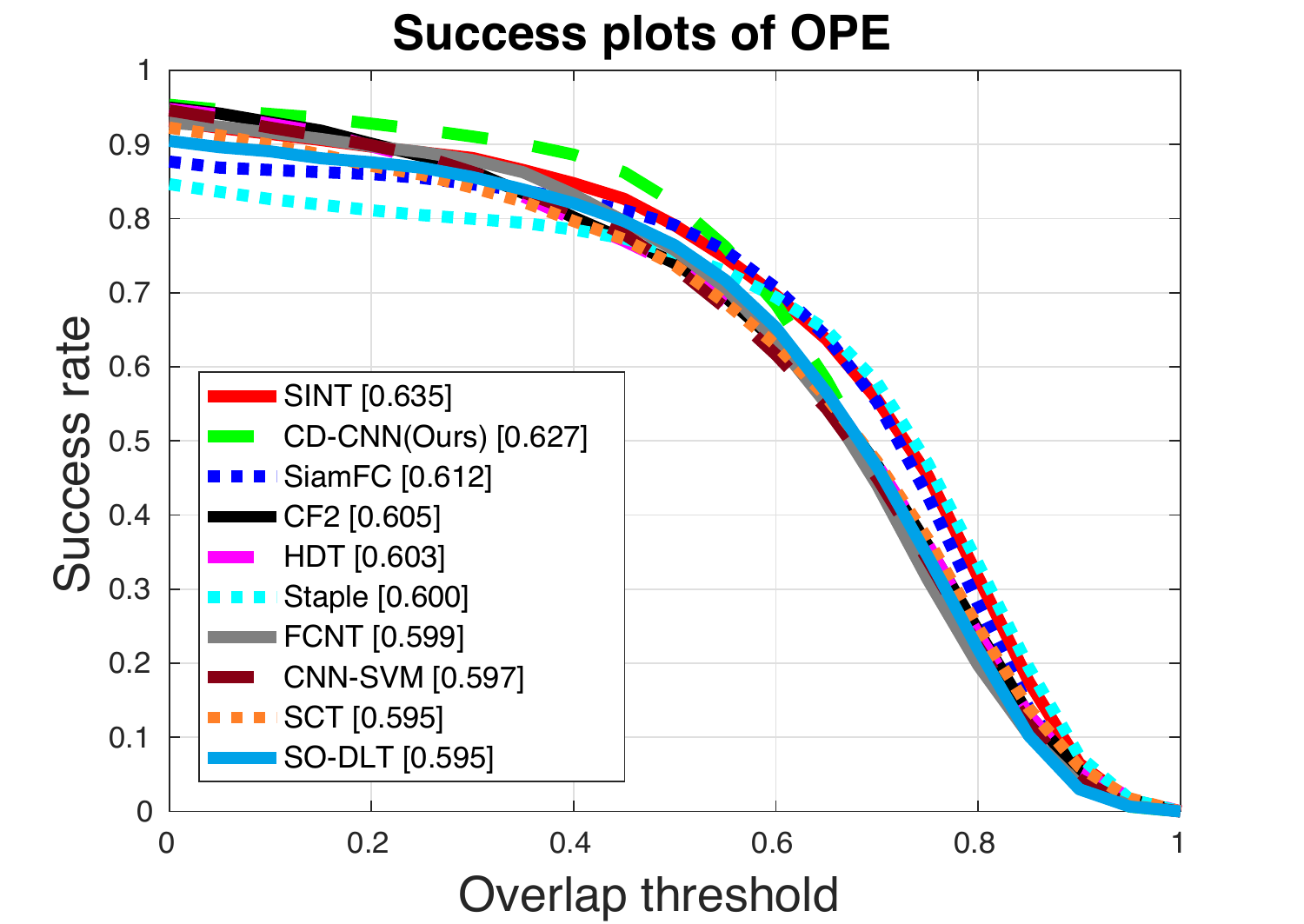}
	\end{varwidth}
	\caption{Overall performance comparison on OTB2013.}
	\label{fig:all_comp_OTB2013}
\end{figure}

\begin{figure}[ht]
	\centering
	\begin{varwidth}[ht]{\textwidth}
		\includegraphics[width=0.23\linewidth]{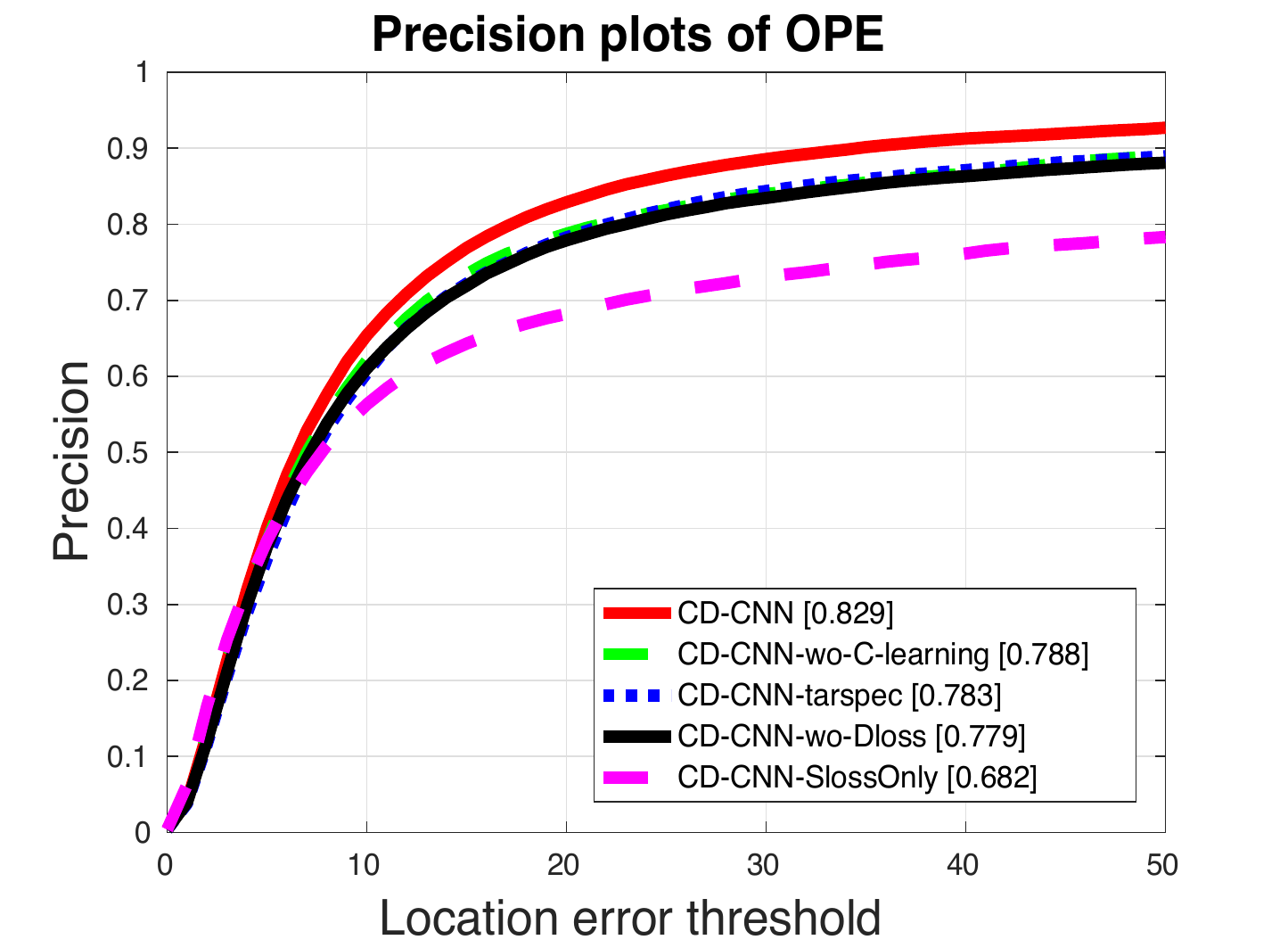}
	\end{varwidth}
	\begin{varwidth}[ht]{\textwidth}
		\includegraphics[width=0.23\linewidth]{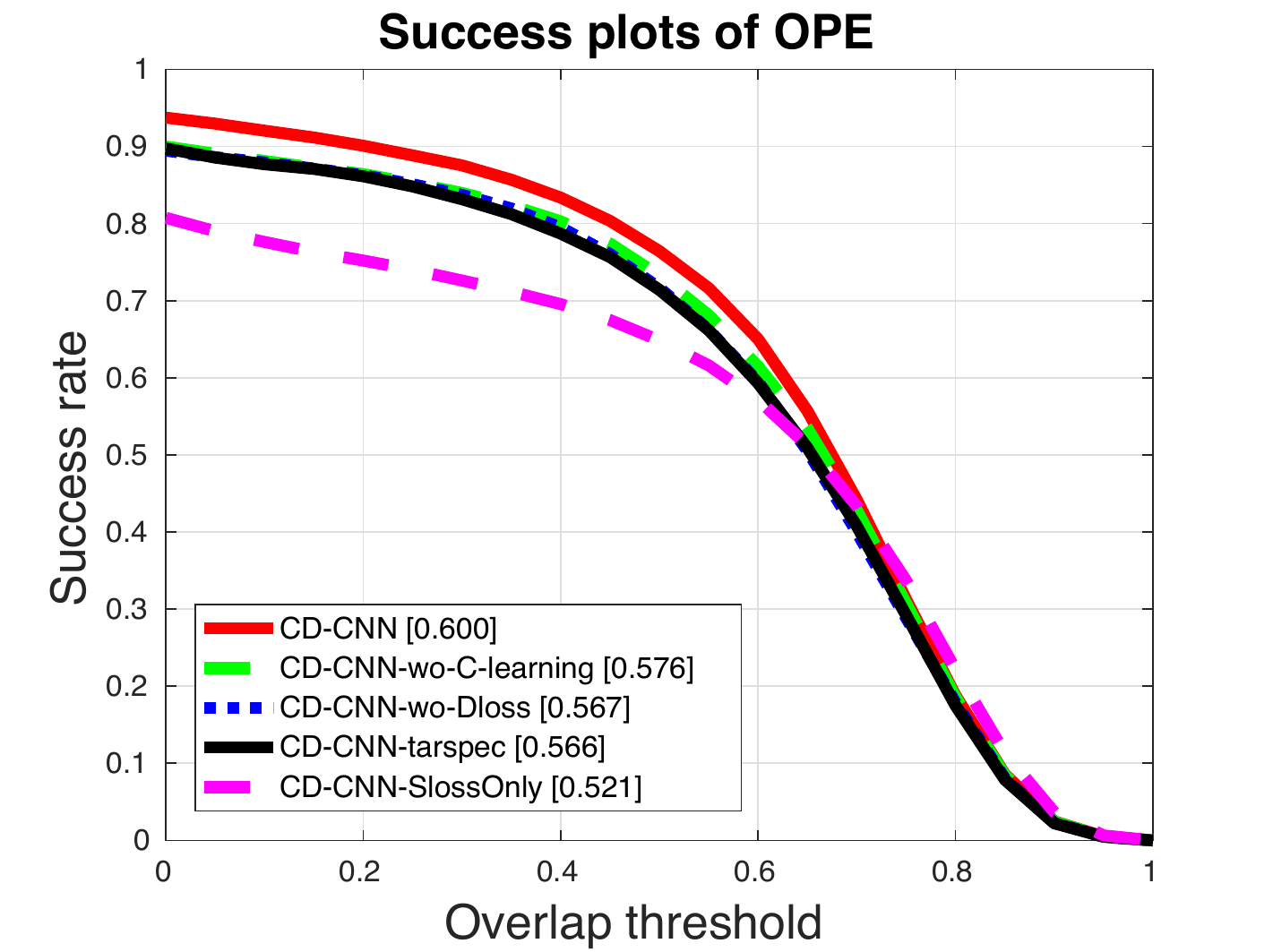}
	\end{varwidth}
	\caption{Internal comparison on OTB2015}
	\label{fig:internal_comp_OTB2015}
\end{figure}

\noindent {\bf Quantitative comparison:}
We employ the one-pass evaluation (OPE) to compare CD-CNN with 15 state-of-the-art trackers including SINT \citep{tao2016siamese}, SiamFC \citep{bertinetto2016fully}, CSR-DCF \citep{Lukezic_CVPR_2017}, CF2 \citep{ma2015hierarchical}, HDT \citep{qi2016hedged}, Staple \citep{bertinetto2015staple:}, FCNT \citep{wang2015visual}, CNN-SVM \citep{hong2015online}, SCT \citep{choi2016visual}, SO-DLT \citep{wang2015transferring}, DLSSVM \citep{ning2016object}, SAMF \citep{li2014a}, MEEM \citep{zhang2014meem:}, DSST \citep{Danelljan785778} and KCF \citep{6870486}. 

Figure \ref{fig:all_comp_OTB100} and Figure \ref{fig:all_comp_OTB2013} illustrate the overall quantitative performance comparison on OTB2015 and OTB2013, respectively. Our tracker achieves 0.600 AUC value on OTB2015 (ranking 1st) and 0.627 on OTB2013 (ranking 2nd), outperforms most of the state-of-the-art trackers in both metrics and demonstrates very promising tracking performance. This validates the effectiveness of introducing temporal appearance continuity and object-centroid into tracking.

\noindent{\bf Qualitative comparison:}
We show qualitative comparison on some typical challenging sequences including CarScale, Ironman and Doll. These sequences cover various challenging conditions involving severe occlusion, fast motion and scaling. As is shown in Figure \ref{fig:qual_perf}, our tracker exhibits surprisingly impressive ability to handle these conditions. 
Especially, in the CarScale, when the occlusion occurs between \#157 and \#180, our tracker can still track the car. 
In \#239, only CD-CNN can successfully track the target with largest IoU in face of fast motion. 
In \#112 of the Ironman, when the head gets out of the view, only CD-CNN can successfully estimate the target position. 
These phenomena benefit from robust features learned from temporal appearance continuity.
In \#2940 of the Doll, SO-DLT yields drifting to the man's face, which is partly due to its coarse inverse mapping. In \#3725, only CD-CNN can successfully handle such rapid scaling, which benefits from its sensitivity to objectness and the relative position of the target in a patch, \ie~object-centroid. 
More analyses of the qualitative performance can be found in Appendix D.

\noindent{\bf Ablation study:}
To empirically validate the effectiveness and necessity of Eqn.~\eqref{eq:c_loss}\eqref{eq:d_loss}\eqref{eq:s_loss} as basic components of the overall loss, we implement and test several variants of CD-CNN, shown in Table~\ref{tab:variants}.
As illustrated in Figure \ref{fig:internal_comp_OTB2015}, on both benchmarks (shown in Appendix C), all the variants above do not perform so well as our original CD-CNN, which validates the formulations of temporal appearance continuity, object-centroid discrimination and the optimization objective in Eqn.~\eqref{eq:min_total_loss} in the offline training stage. 

\begin{table}[t]
\begin{center}
\caption{CD-CNN variants for ablation study. fc-ot? denotes whether fc3, fc4 and fc5 are trained offline or not.} \label{tab:variants}
\small
\begin{tabular} {c|ccccc}
Model Name & $\mathcal L^C$ & $\mathcal L^P$ & $\mathcal L^D$ & $\mathcal L^S$ & fc-ot?\\
\hline
CD-CNN & $\checkmark$ & $-$ & $\checkmark$ & $\checkmark$ & $\checkmark$\\   
CD-CNN-wo-C-learning & $-$ & $\checkmark$ & $\checkmark$ & $\checkmark$ & $\checkmark$\\
CD-CNN-wo-Dloss & $\checkmark$ & $-$ & $-$ & $\checkmark$ & $\checkmark$\\   
CD-CNN-SlossOnly & $-$ & $-$ & $-$ & $\checkmark$ & $\checkmark$\\
CD-CNN-tarspec & $\checkmark$ & $-$ & $\checkmark$ & $\checkmark$ & $-$\\
\hline
\end{tabular}
\end{center}
\end{table}

\noindent{\bf Objectness comparison:}
To the best of our knowledge, SO-DLT was the first and the only one that introduced objectness into visual tracking. Empirically, CD-CNN performs favorably against SO-DLT with a large margin in both metrics. To see the advantage of object-centroid over objectness, we compare CD-CNN-wo-C-learning (0.576 AUC value) with SO-DLT (0.561 AUC value). This again demonstrates the superiority of object-centroid discrimination learning. 

\section{Conclusions}
In this paper, we have proposed a novel deep model for visual object tracking, CD-CNN, which simultaneously characterizes two fundamental properties in visual tracking, temporal appearance continuity and object-centroid. 
Mathematically, we have proved, by introducing temporal appearance continuity into tracking, the upper bound of target appearance representation error can be sufficiently small with high probability. 
Empirically, we have verified the effectiveness and necessity of temporal appearance continuity transferring and object-centroid discrimination learning. Extensive experimental results have demonstrated the competitive tracking performance of our method in comparison with state-of-the-art trackers. 
\\

\noindent{\bf Acknowledgement}
This work is supported in part by 973 Program under the contract No. 2015CB351802, and Natural Science Foundation of China (NSFC): 61390515, 61390511, 61572465 and 61650202.

\small
\bibliographystyle{IEEEbib}
\bibliography{ref}

\onecolumn
\appendix
\section*{Appendices}
\addcontentsline{toc}{section}{Appendices}


\renewcommand{\thesubsection}{\Alph{subsection}}

\subsection{Algorithm 1}
\begin{algorithm*}[h]
\caption{CD-CNN online tracking algorithm}
\label{alg:online_tracking}
\begin{algorithmic}[1]
\REQUIRE ~~\\
PVANET pool5 feature extractor;\\
pre-trained learning weights $W=\{W_1, ..., W_5\}$;\\
initial target patch $P_*^{(1)}$ with its bounding box $B_*^{(1)}$;
\ENSURE ~~\\
Predicted target patches ${\{\hat{P}_*^{(t)}\}_{t=2}^T}$ with bounding boxes $\{\hat{B}_*^{(t)}\}_{t=2}^T$;
\STATE Draw positive samples $\{P_j^{(1)}\}$, $\{P_k^{(1)}\}$ and negative samples $\{N_l^{(1)}\}$;
\STATE Update $\{W_1, ..., W_5\}$ using $\{P_j^{(1)}\}$, $\{P_k^{(1)}\}$, $\{N_l^{(1)}\}$;
\STATE $t=2$;
\REPEAT 
\STATE Draw $m$ candidates $\{P_j^{(t)}\}_{j=1}^m$, with corresponding bounding boxes $\{B_j^{(t)}\}$;
\STATE Forward $m$ candidates through CD-CNN;
\STATE Select the top five candidates $\{P_{(1)}^{(t)}, ..., P_{(5)}^{(t)}\}$ with highest object-centroid scores $p(\cdot)$;
\STATE Determine $\hat{P}_*^{(t)}$ by averaging bounding boxes $\{B_{(1)}^{(t)}, ..., B_{(5)}^{(t)}\}$;
\IF{$t$ mod 5 == 0 and $p(\hat{P}_*^{(t)})>0.95$}
\STATE Draw positive samples $\{P_j^{(t)}\}$, $\{P_k^{(t)}\}$ and negative samples $\{N_l^{(t)}\}$;
\STATE Update $\{W_1, ..., W_5\}$ using $\{P_j^{(t)}\}$, $\{P_k^{(t)}\}$, $\{N_l^{(t)}\}$;
\ENDIF
\STATE $t=t+1$;
\UNTIL{end of sequence}
\end{algorithmic}
\end{algorithm*}

\clearpage

\subsection{Proof of Theorem 1}
The target appearance representation error incurred by our tracking method can be defined as the Euclidean distance $\mathcal{E} := {\lVert \Phi({\hat P}_*^{(t+1)})-\Phi(P_*^{(t+1)}) \rVert}_2$, where $P_*^{(t)}$ is the ground-truth and ${\hat P}_*^{(t)}$ is the predicted target patch in the $t$th frame.

\begin{thm}[Upper Bound of Target Appearance Representation Error]
With probability no less than $1-\rho$, the target prediction error $\mathcal{E}$ is upper-bounded by $\sum_j{\sqrt{\hat{\mathcal{L}_j^{C}}}}/m + n(\delta + K\Delta t)$, for any $\delta > \sqrt{{n \over m} max_i{\mathrm{Var}\left(\Phi_i\right)}}$, where $\hat{\mathcal{L}_j^{C}} = {\lVert \Phi(\hat{P}_*^{(t+1)}) - \Phi(P_j^{(t)}) \rVert}_2^2$ is the estimated temporal appearance continuity loss for the predicted target in the $(t+1)$th frame with respect to $P_j^{(t)}$ and $\rho = n\max_i{\mathrm{Var}(\Phi_i)}/{m\delta^2}$.
\end{thm}

\begin{proof}
The target appearance representation error incurred at each time slot is given by
\begin{equation}
\begin{split}
\mathcal{E}
& = {\left\lVert \Phi\left({\hat P}_*^{(t+1)}\right)-\Phi\left(P_*^{(t+1)}\right) \right\rVert}_2 \\
& = {\left\lVert \Phi\left({\hat P}_*^{(t+1)}\right)-\Phi\left(P_*^{(t)}\right)+\Phi\left(P_*^{(t)}\right)-\Phi\left(P_*^{(t+1)}\right) \right\rVert}_2 \\
& \le {\left\lVert \Phi\left({\hat P}_*^{(t+1)}\right)-\Phi\left(P_*^{(t)}\right) \right\rVert}_2 + \epsilon \\
& = {\left\lVert \Phi\left({\hat P}_*^{(t+1)}\right)-\frac{1}{m}\displaystyle\sum_{j=1}^m\Phi\left(P_j^{(t)}\right)+\frac{1}{m}\displaystyle\sum_{j=1}^m\Phi\left(P_j^{(t)}\right)-\Phi\left(P_*^{(t)}\right) \right\rVert}_2 + \epsilon \\
& \le {\left\lVert \Phi\left({\hat P}_*^{(t+1)}\right)-\frac{1}{m}\displaystyle\sum_{j=1}^m\Phi\left(P_j^{(t)}\right) \right\rVert}_2 +  {\left\lVert \frac{1}{m}\displaystyle\sum_{j=1}^m\Phi\left(P_j^{(t)}\right)-\Phi\left(P_*^{(t)}\right) \right\rVert}_2 + \epsilon \\
& = {\left\lVert \Phi\left({\hat P}_*^{(t+1)}\right)-\frac{1}{m}\displaystyle\sum_{j=1}^m\Phi\left(P_j^{(t)}\right) \right\rVert}_2 + {\left\lVert \overline{\Phi\left(P^{(t)}\right)} - \Phi\left(P_*^{(t)}\right) \right\rVert}_2 + \epsilon \\
& \le {\left\lVert \Phi\left({\hat P}_*^{(t+1)}\right)-\frac{1}{m}\displaystyle\sum_{j=1}^m\Phi\left(P_j^{(t)}\right) \right\rVert}_2 + {\left\lVert \overline{\Phi\left(P^{(t)}\right)} - \Phi\left(P_*^{(t)}\right) \right\rVert}_1 + \epsilon \\
& = {\left\lVert \Phi\left({\hat P}_*^{(t+1)}\right)-\frac{1}{m}\displaystyle\sum_{j=1}^m\Phi\left(P_j^{(t)}\right) \right\rVert}_2 + \displaystyle\sum_{i=1}^n{\left\vert \overline{\Phi_i\left(P^{(t)}\right)}-\Phi_i\left(P_*^{(t)}\right) \right\vert} + \epsilon
\end{split}
\end{equation}
where $P_j^{(t)}$ is the $j$th positive samples drawn around $P_*^{(t)}$ and $\overline{\Phi_i\left(P^{(t)}\right)}$ denotes the arithmetic mean. Mathematically, it is assumed that, in the feature space $\chi$, the random vector $\Phi$, obeys some unknown distribution $\mathbb{P}(\varphi)$, whose expectation is given by
\begin{equation}
\mathbb{E}\left[\Phi\left(P_j^{(t)}\right)\right]=\int_{\chi} \varphi\, d\mathbb{P}(\varphi) =  \Phi\left(P_*^{(t)}\right)
\end{equation}
By Chebyshev inequality and Inclusion-exclusion Principle,
\begin{equation}
\begin{split}
\mathbb{P}\left(\bigcap_{i=1}^n {\left\vert \overline{\Phi_i\left(P^{(t)}\right)} - \Phi_i\left(P_*^{(t)}\right) \right\vert < \delta}\right)
& \ge 1 - \displaystyle\sum_{i=1}^n {\mathbb{P}\left(\left\vert \overline{\Phi_i\left(P^{(t)}\right)} - \Phi_i\left(P_*^{(t)}\right) \right\vert \ge \delta\right)} \\
& \ge 1 - \frac{1}{m\delta^2}\displaystyle\sum_{i=1}^n{\mathrm{Var}(\Phi_i)} \\
& \ge 1 - \frac{n}{m\delta^2}\max_i{\mathrm{Var}(\Phi_i)}
\end{split}
\end{equation}
Therefore, with the lower-bounded probability above, the target appearance representation error incurred at each time slot is upper-bounded by
\begin{equation}
\begin{split}
\mathcal{E}
& \le {\left\lVert \Phi\left(\hat{P}_*^{(t+1)}\right) - \frac{1}{m}\displaystyle\sum_{j=1}^{m}{\Phi\left(P_j^{(t)}\right)} \right\rVert}_2 + \displaystyle\sum_{i=1}^{n}{\left\vert \overline{\Phi_i\left(P^{(t)}\right)} - \Phi_i\left(P_*^{(t)}\right)\right\vert} + {\epsilon} \\
& \le \frac{1}{m}\displaystyle\sum_j{\left\lVert \Phi\left(\hat{P}_*^{(t+1)}\right) - \Phi\left(P_j^{(t)}\right) \right\rVert}_2 + n\delta + \epsilon \\
& = {1 \over m} \displaystyle\sum_j{\sqrt{{\left\lVert \Phi\left(\hat{P}_*^{(t+1)}\right) - \Phi\left(P_j^{(t)}\right) \right\rVert}_2^2}}+n\delta+\epsilon \\
& \le \frac{1}{m}\displaystyle\sum_j{{\left(\hat{\mathcal{L}_j^{C}}\right)}^{1 \over 2}} + n\left(\delta + K\Delta t\right)
\end{split}
\end{equation}
\end{proof}

\subsection{Quantitive Comparison}
\subsubsection{OTB2015 Comparison}
\begin{figure}[h]
	\centering
	\begin{varwidth}[h]{\textwidth}
		\includegraphics[width=0.46\linewidth]{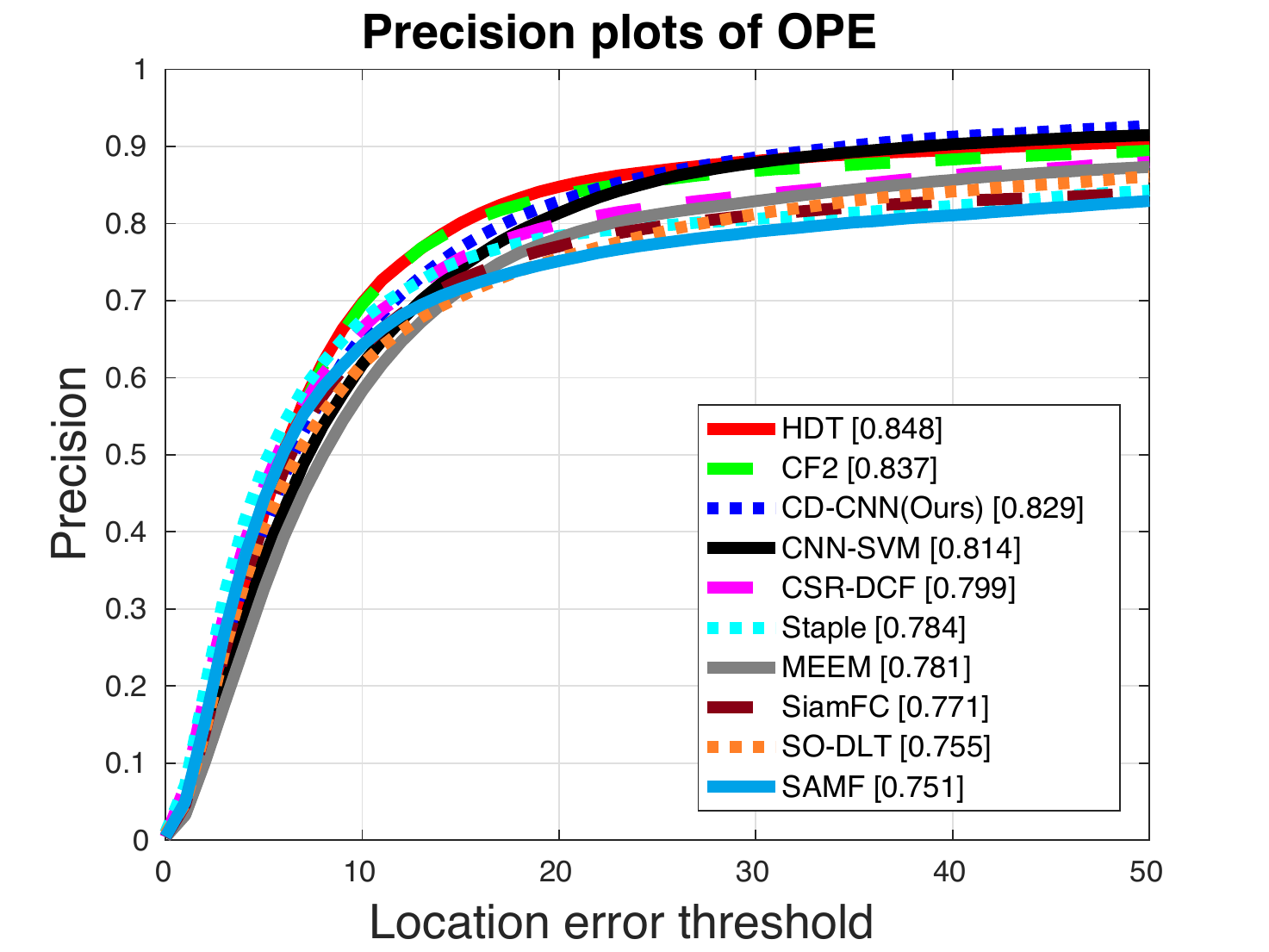}
	\end{varwidth}
	\qquad
	\begin{varwidth}[h]{\textwidth}
		\includegraphics[width=0.46\linewidth]{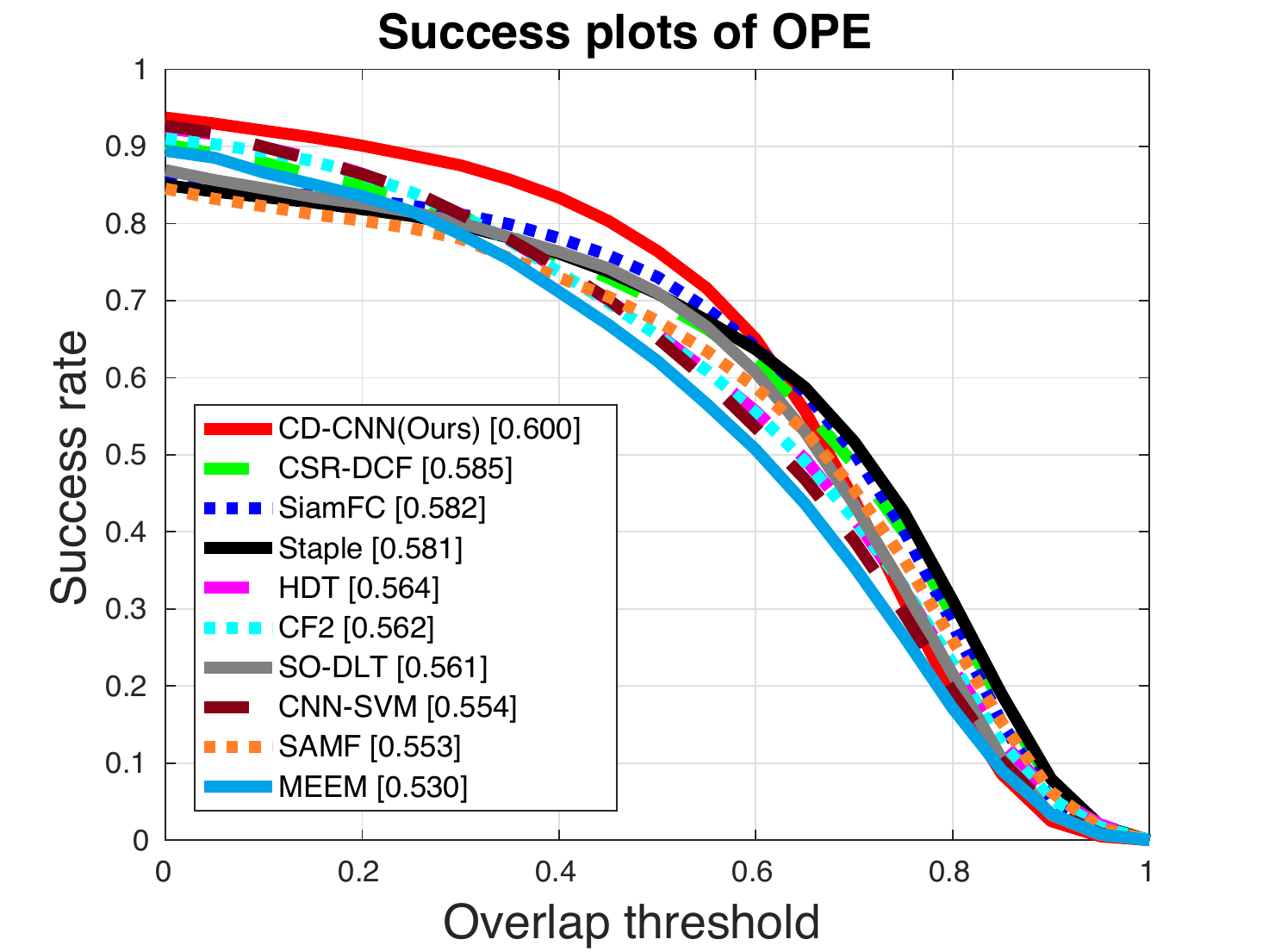}
	\end{varwidth}
	\caption{Precision plots and success plots for the overall performance comparison on OTB2015.}
	\label{fig:all_comp_OTB2015}
\end{figure}

Figure \ref{fig:all_comp_OTB2015} and Figure \ref{fig:internal_comp_OTB2015} show the overall performance comparison with state-of-the-art trackers and the internal comparison, respectively, on the OTB2015 dataset.
\begin{figure}[h]
	\centering
	\begin{varwidth}[h]{\textwidth}
		\includegraphics[width=0.46\linewidth]{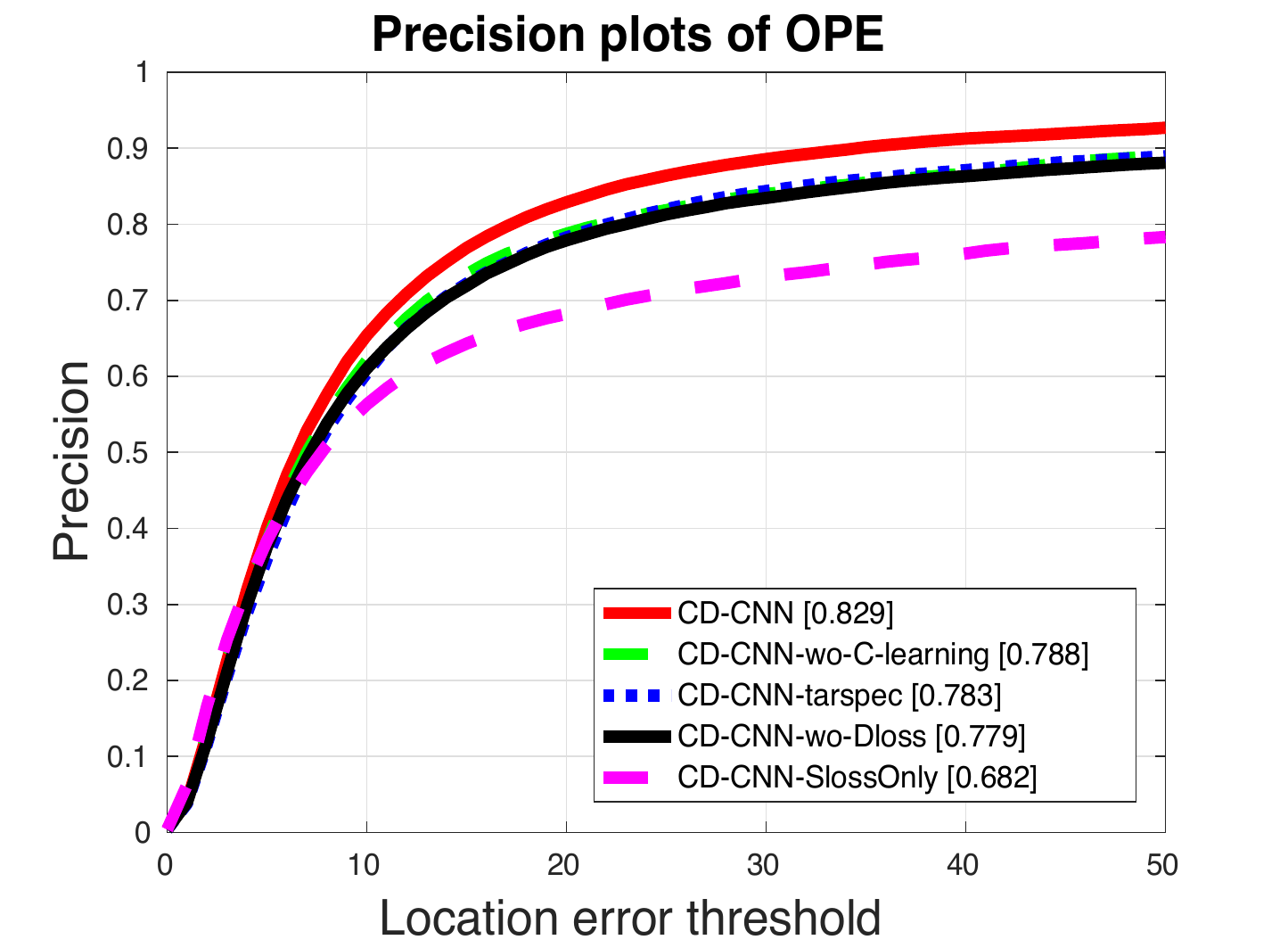}
	\end{varwidth}
	\qquad
	\begin{varwidth}[h]{\textwidth}
		\includegraphics[width=0.46\linewidth]{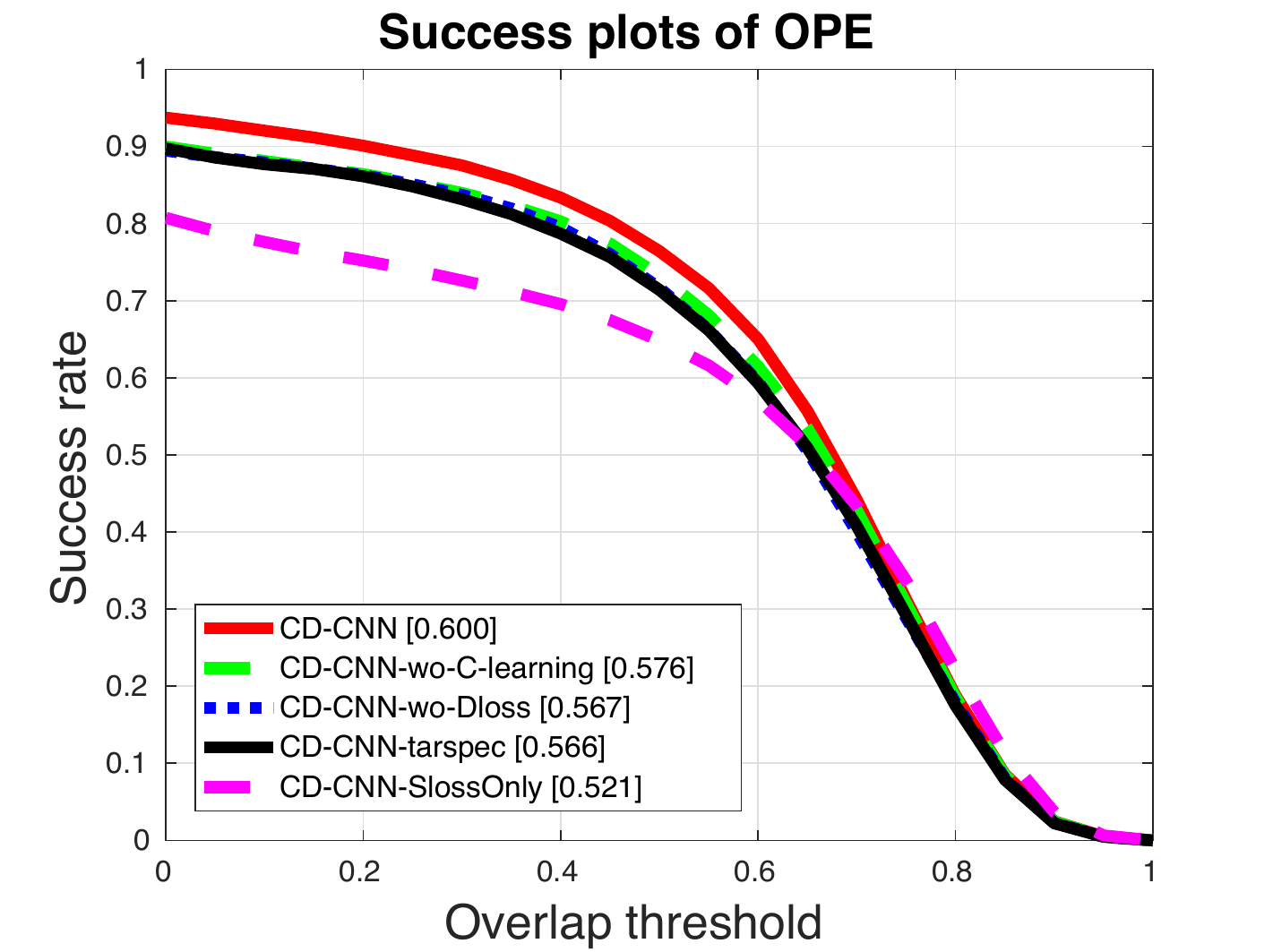}
	\end{varwidth}
	\caption{Precision plots and success plots of OPE for the internal comparisons on OTB2015.}
	\label{fig:internal_comp_OTB2015}
\end{figure}
\subsubsection{OTB2013 Comparison}
\begin{figure}[h]
	\centering
	\begin{varwidth}[h]{\textwidth}
		\includegraphics[width=0.46\linewidth]{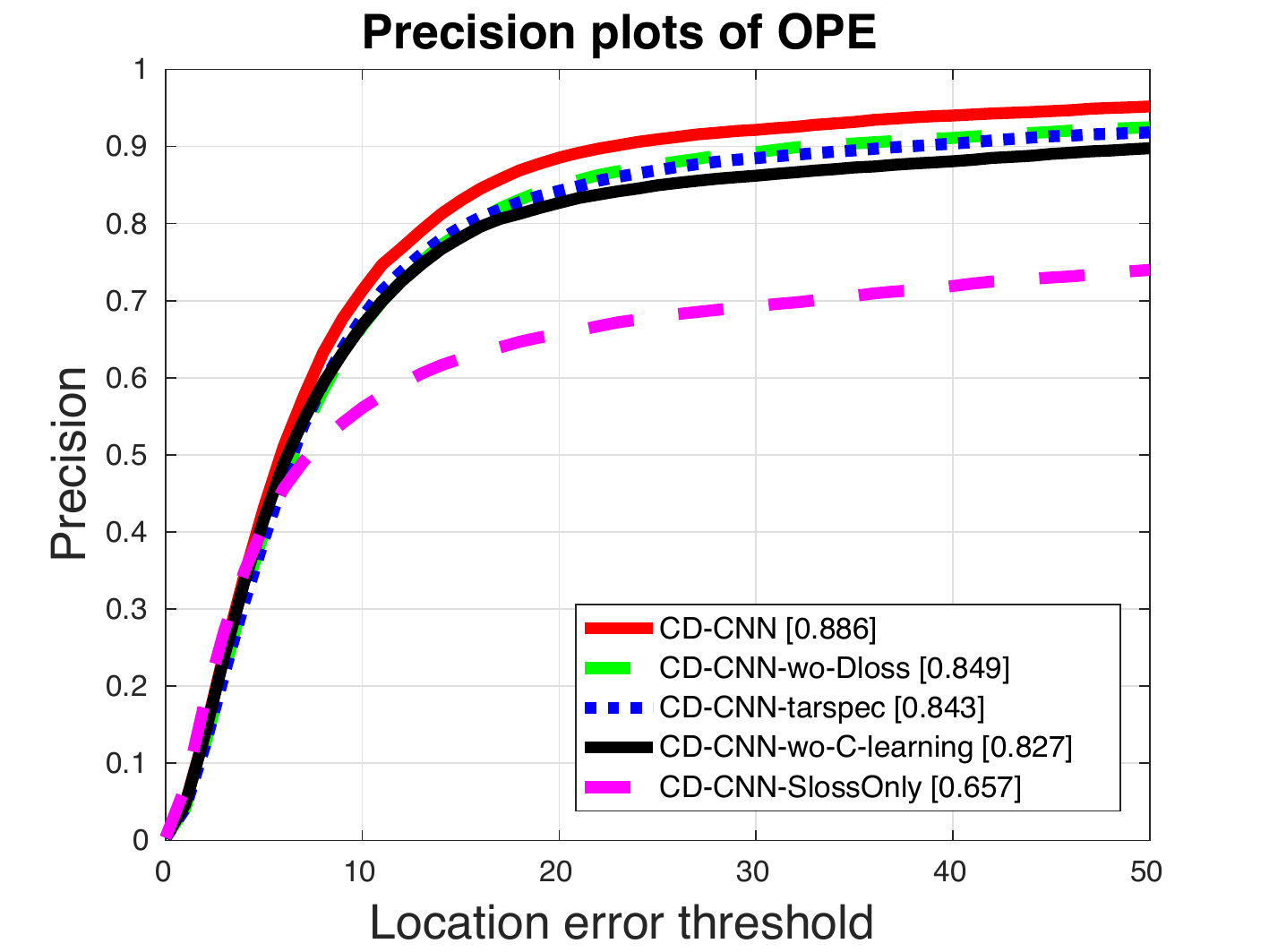}
	\end{varwidth}
	\qquad
	\begin{varwidth}[h]{\textwidth}
		\includegraphics[width=0.46\linewidth]{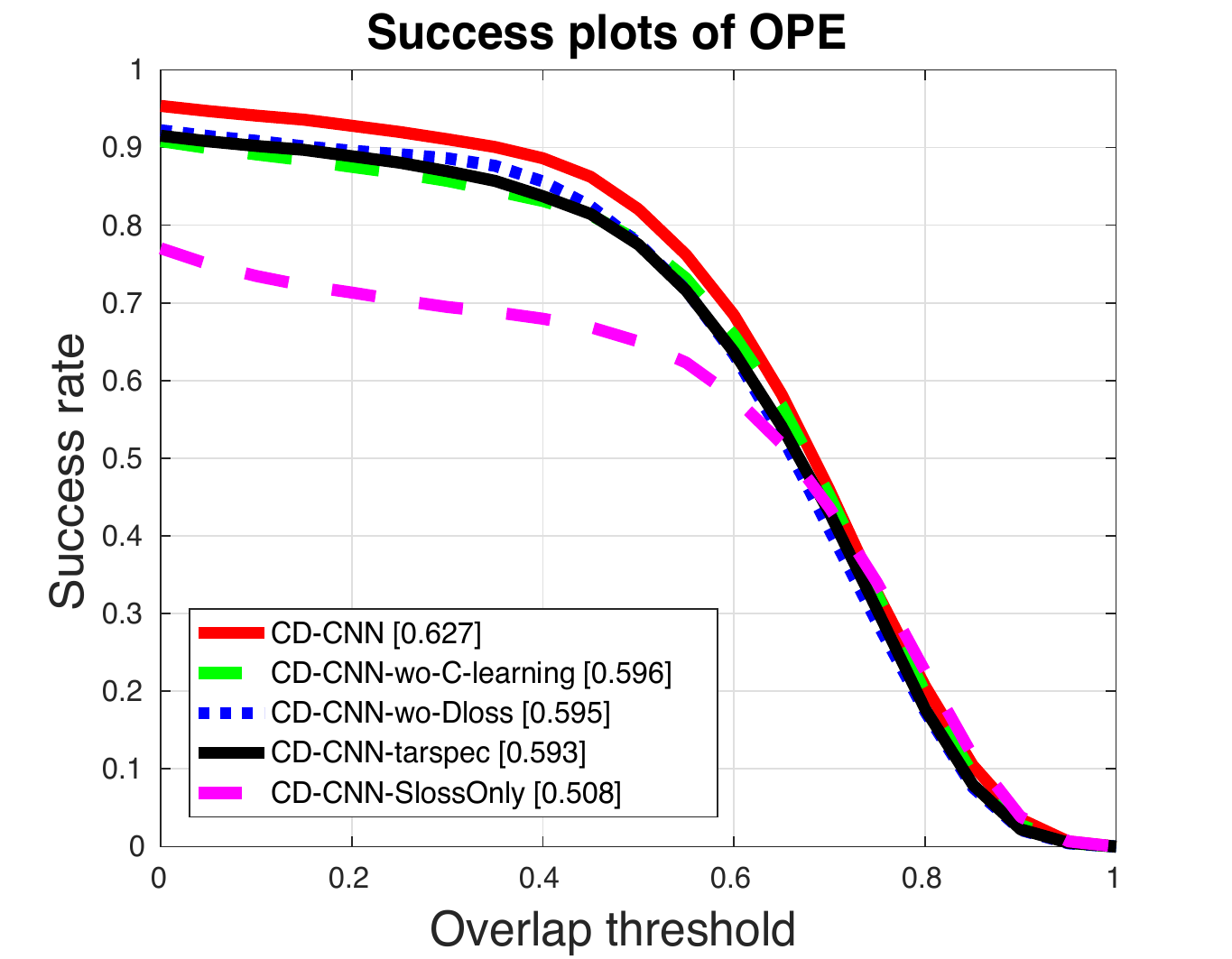}
	\end{varwidth}
	\caption{Precision plots and success plots of OPE for the internal comparisons on OTB2013.}
	\label{fig:internal_comp_OTB2013}
\end{figure}

Figure \ref{fig:all_comp_OTB2013} and Figure \ref{fig:internal_comp_OTB2013} show the overall performance comparison with state-of-the-art trackers and the internal comparison, respectively, on the OTB2013 dataset.
\begin{figure}[h]
	\centering
	\begin{varwidth}[h]{\textwidth}
		\includegraphics[width=0.46\linewidth]{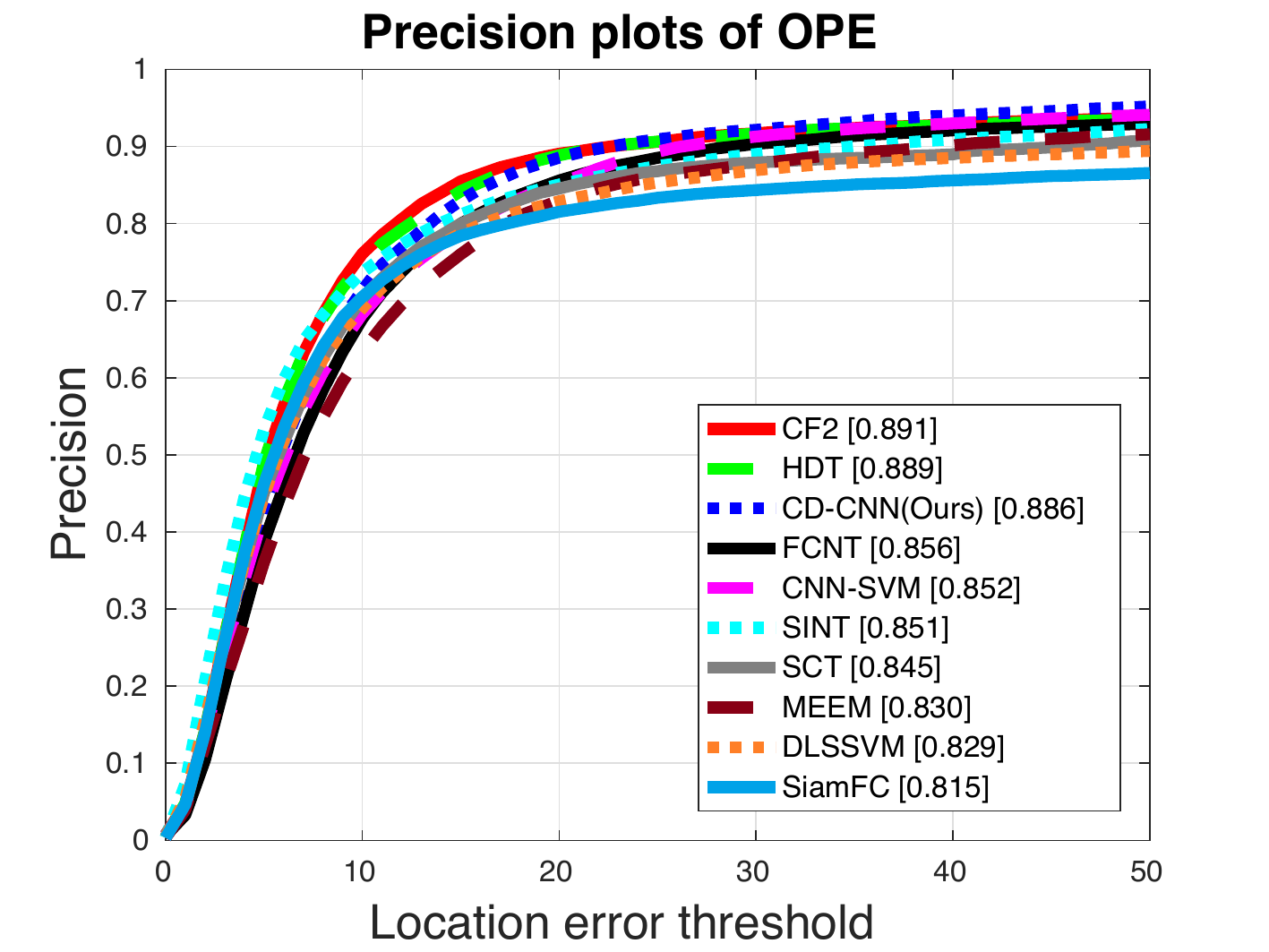}
	\end{varwidth}
	\qquad
	\begin{varwidth}[h]{\textwidth}
		\includegraphics[width=0.46\linewidth]{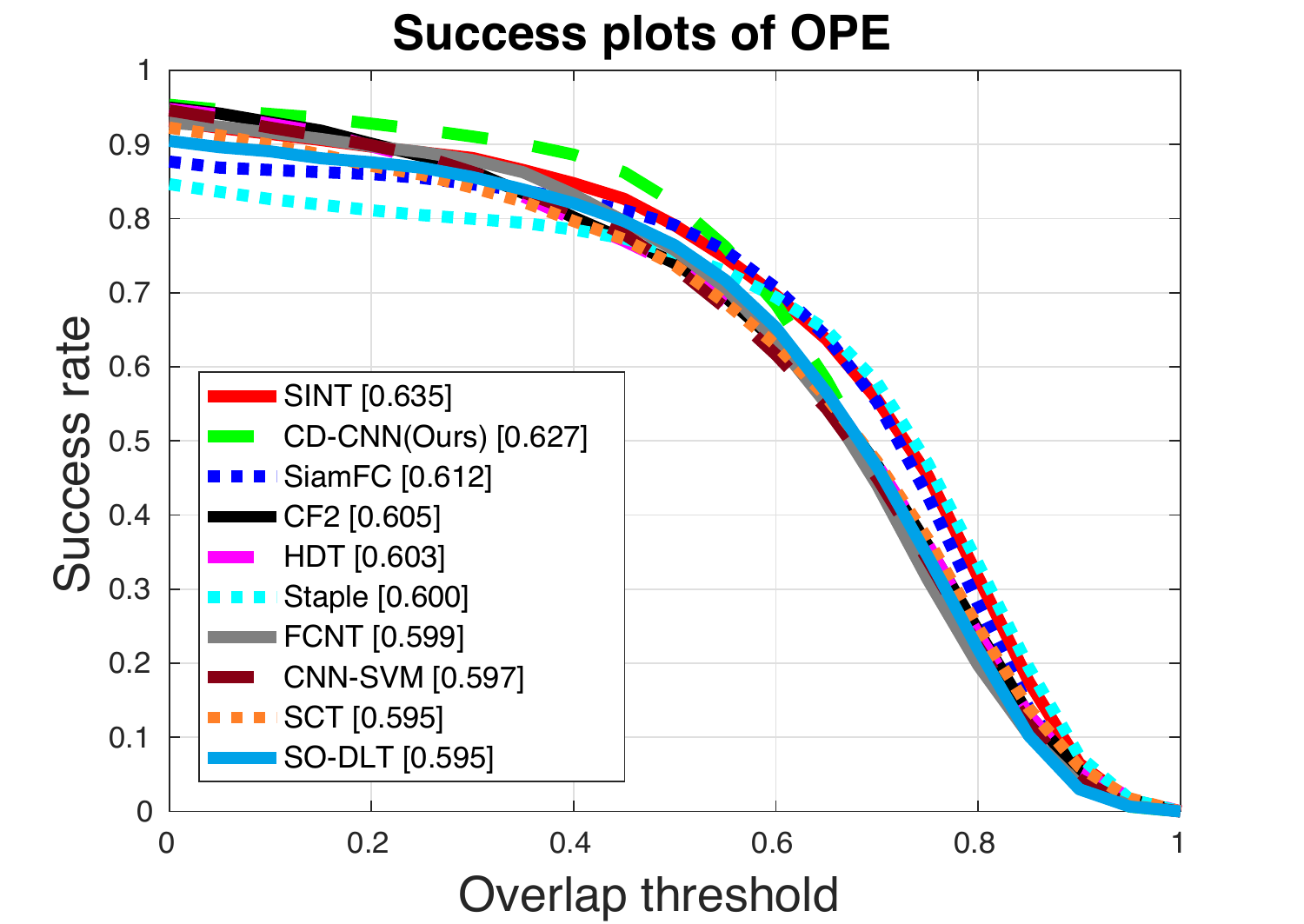}
	\end{varwidth}
	\caption{Precision plots and success plots for the overall performance comparison on OTB2013.}
	\label{fig:all_comp_OTB2013}
\end{figure}

\subsection{Qualitative Comparison}
Figure \ref{fig:qual_perf} illustrates the qualitative performance of our tracker on some challenging sequences, compared with state-of-the-art trackers including SINT, SiamFC, CF2, HDT and SO-DLT. 

In the CarScale sequence, when the occlusion occurs between \#157 and \#180, our tracker can still track the car. In \#239 of CarScale, only CD-CNN can successfully track the target with the largest IoU. 

In \#2940 of the Doll sequence, SO-DLT yields drifting to the man's face. This might be due to its inaccurate inverse mapping. In \#3725, only CD-CNN can successfully handle the rapid scale variation, which benefits from its sensitivity to objectness and the relative position of the target in a patch, that is, object-centroid. In \#3769, CD-CNN outperforms others by successfully tracking the blurred target, as a result of its temporal appearance continuity transferring.

In \#562 and \#574 of the Woman sequence, when the camera zooms in, CD-CNN is the only tracker that can successfully handle the rapid scale variation.

In the Football sequence, SINT, which focuses on learning an implicit matching function, fails to track the target when a similar object appears nearby. Apparently, its lack of discriminability causes the drifting. 

In \#112 of the Ironman sequence, when the head gets out of the view, only CD-CNN can successfully estimate the target position. This benefits from the temporal appearance continuity learning. Again, in \#117 and \#141, our CD-CNN apparently outperforms other trackers while the similarity matching based tracker including SINT, SiamFC and CF2 drifts into similar background. 

In the end of the Skiing sequence, only CD-CNN yields a tight bounding box for the target. This is again attributed to the object-centroid discrimination of our model.

\begin{figure}[t]
\centering
\includegraphics[width=\linewidth]{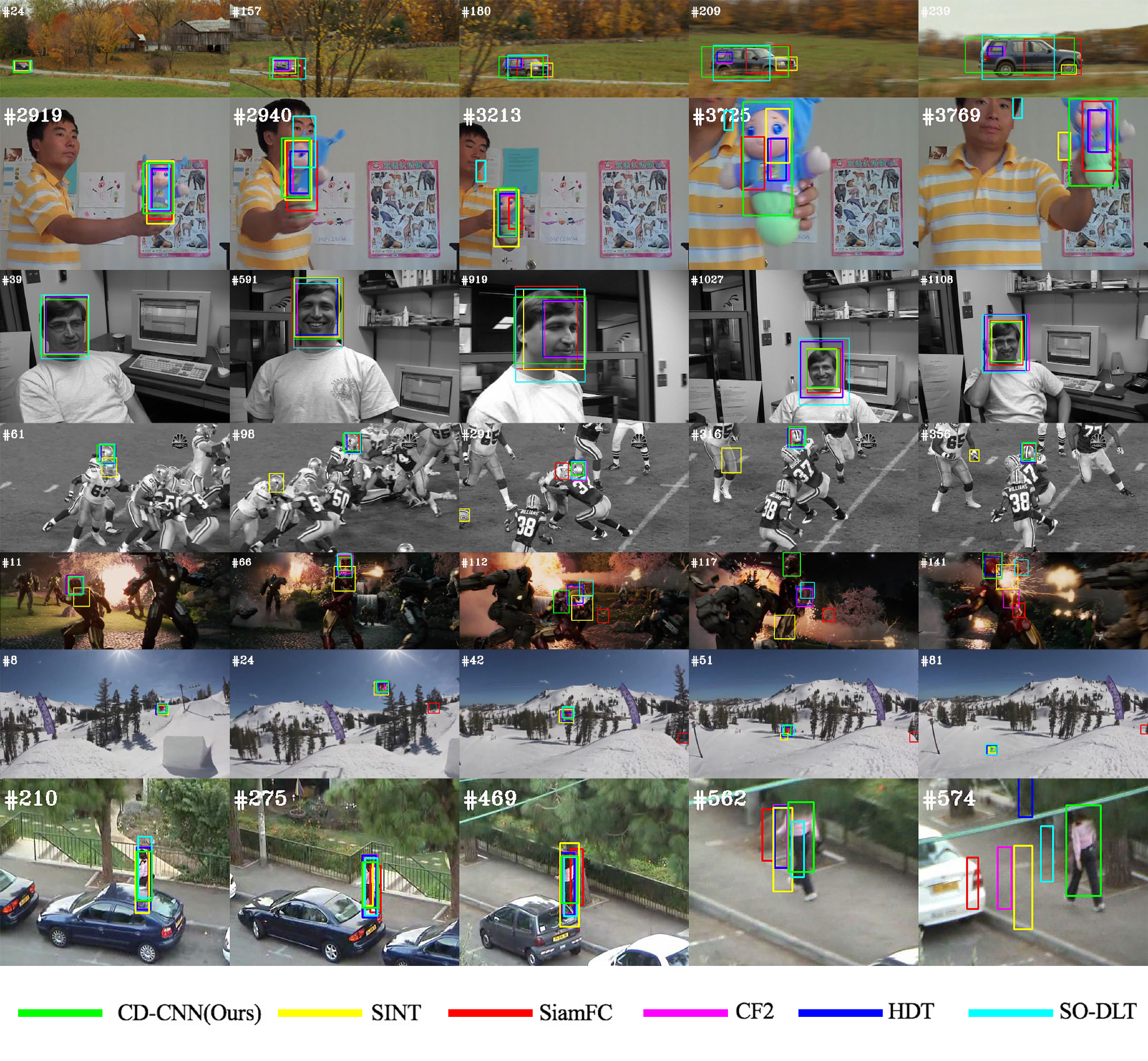}
\caption{Qualitative comparisons on some challenging sequences including CarScale, Doll, Dudek, Football, Ironman, Skiing and Woman.}
\label{fig:qual_perf}
\end{figure}

\end{document}